\documentclass{article}
\usepackage{xcolor}
\usepackage[preprint]{corl_2026}
\usepackage{graphicx,booktabs,amsmath,microtype,longtable,array,placeins,etoolbox}
\usepackage[T1]{fontenc}
\graphicspath{{revision_assets/}{figures/}}
\definecolor{mover}{HTML}{147D92}
\definecolor{reference}{HTML}{C46635}
\title{Same Goal, Different Words:\\Spatial Instruction Sensitivity in World-Action Models}
\author{Ali-Adeeb Abbas$^{1,2}$, Esmaeil Seraj$^2$, Behrad Toghi$^2$, and Ta\c{s}k{\i}n Pad{\i}r$^1$\thanks{$^1$RIVeR Laboratory, Northeastern University, Boston, MA, USA. $^2$General Motors, Mountain View, CA, USA. Correspondence: \texttt{abbas.a@northeastern.edu}.\newline Ta\c{s}k{\i}n Pad{\i}r holds concurrent appointments as a Professor of Electrical and Computer Engineering at Northeastern University and as an Amazon Scholar. This paper describes work performed at Northeastern University and is not associated with Amazon.}}
\hypersetup{pdftitle={Same Goal, Different Words: Spatial Instruction Sensitivity in World-Action Models},pdfsubject={arXiv preprint},pdfauthor={Ali-Adeeb Abbas, Esmaeil Seraj, Behrad Toghi, Ta\c{s}k{\i}n Pad{\i}r}}

\makeatletter
\patchcmd{\@maketitle}{\vskip 0.1in}{\vskip 0pt}{}{\PackageError{preprint-layout}{Title top spacing patch failed}{}}
\patchcmd{\@maketitle}{\vskip 0.3in \@minus 0.1in}{\vskip 0.1in}{}{\PackageError{preprint-layout}{Title bottom spacing patch failed}{}}
\patchcmd{\@maketitle}{\bf\rule{\z@}{24\p@}\@author}{\bf\rule{\z@}{12\p@}\@author}{}{\PackageError{preprint-layout}{Author spacing patch failed}{}}
\patchcmd{\@maketitle}{\bf\rule{\z@}{24\p@}\@author}{\bf\rule{\z@}{12\p@}\@author}{}{\PackageError{preprint-layout}{Preprint author spacing patch failed}{}}
\makeatother

\begin{document}
\maketitle
\begin{abstract}
World-action models should preserve the intended manipulation across equivalent descriptions of a spatial goal. We evaluate this consistency in 864 rollouts from three models across four RoboLab scenes, holding initial states and task goals fixed. Describing relations through the reference object rather than the object-to-move lowers the rate of reaching a stable placement by 12.2 percentage points on average. The effect of instruction rephrasing varies across spatial relations, so aggregate scores can hide sensitivity to wording. Standalone tests of the policies' language understanding identify relation reversals and show that adding images can disrupt otherwise correct text-only interpretations. Rollout analysis also distinguishes spatial placement from instruction following: moving the wrong object can satisfy the requested relation, and rates of stable placement fall from 43.2\% at any time to 21.8\% at episode end. These findings motivate a more thorough evaluation of world-action models’ steerability, language understanding, and grounding. Evaluations should test consistency across equivalent instructions, verify that the intended object moved, and check whether the requested arrangement persists to the end of the rollout.

\end{abstract}

\setlength{\parskip}{2.5pt}
\setlength{\intextsep}{8pt plus 2pt minus 2pt}
\setlength{\textfloatsep}{12pt plus 2pt minus 2pt}
\section{Introduction}
World-action models (WAMs) generate robot actions together with predictions of future observations~\citep{kim2026cosmospolicy}. Their usefulness as general-purpose policies depends on turning a user's description into the intended manipulation. For a spatial placement task, this requires identifying both the object to move and its destination. Does the policy respond consistently to equivalent language instructions?

Prior work tests whether policies respond to language in visually similar scenes~\citep{tri2025lbm}, follow changed goals~\citep{fang2026counterfactual}, or withstand paraphrases and irrelevant context~\citep{fei2026liberoplus,kim2026liberopara,moskalenko2026bring}. CAST and SG-WAM seek to improve instruction following through counterfactual supervision and semantic guidance~\citep{glossop2025cast,he2026sgwam}. We focus on a specific consistency requirement: preserving the intended manipulation when the same spatial goal is described from another object's perspective.

We compare three released WAMs under alternative instructions from identical initial scenes. Success under one description does not imply consistency across equivalent descriptions, and opposite effects across spatial relations can cancel in an overall score. Examining the policy rollouts also reveals a limitation of spatial evaluation: the requested relation can become true because the wrong object moved, or disappear under continued control. To isolate language understanding from action generation, we also test the world-action models’ vision-language backbones independently to determine whether the same spatial descriptions lead to different interpretations.

\begin{table}[!t]
\caption{One goal described three ways. The mustard is the \emph{target} to move; the raisin box is the \emph{reference}. Only the instruction changes between runs.}
\label{tab:setup}
\small
\begin{minipage}[c]{.30\linewidth}
\textbf{Initial scene}\par\smallskip
\includegraphics[trim=380bp 85bp 360bp 100bp,clip,angle=90,width=\linewidth]{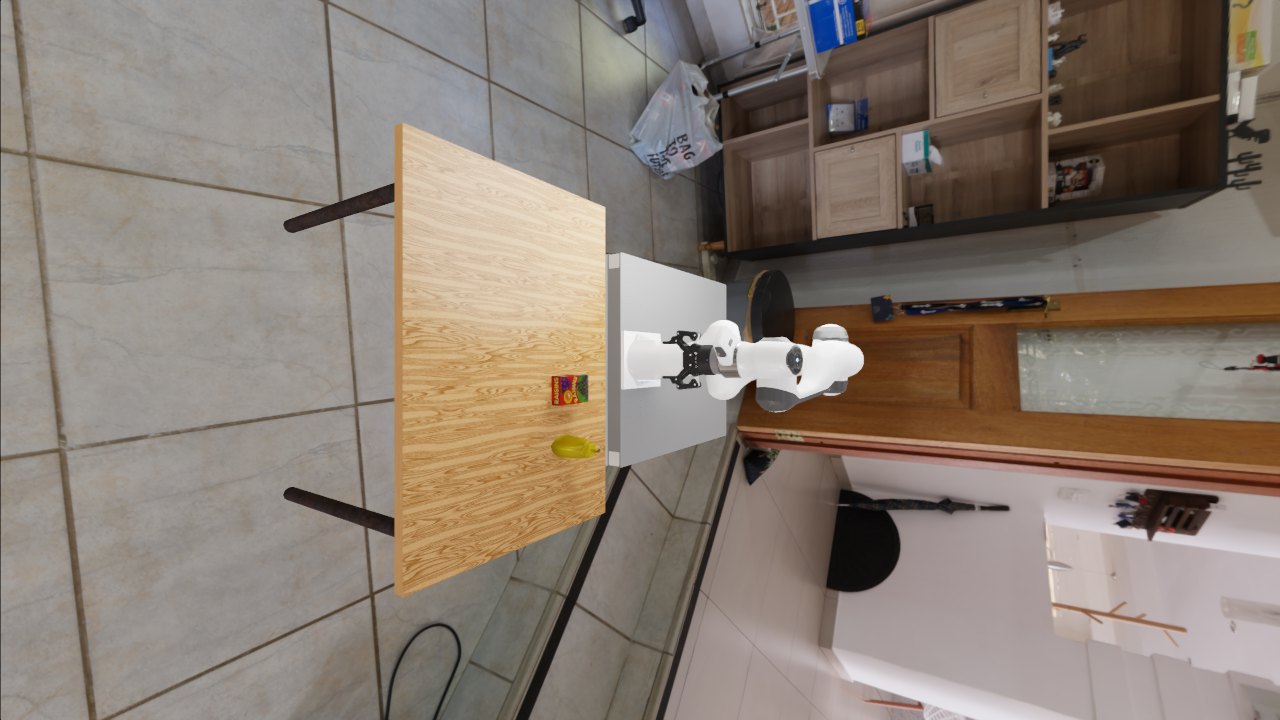}
\end{minipage}\hfill
\begin{minipage}[c]{.66\linewidth}
\textbf{Goal (robot frame):} mustard to the right of the box.\par
TF and RF differ in the subject of the spatial clause. Both still ask the robot to move the mustard.\par\smallskip
\begin{tabular}{@{}p{\linewidth}@{}}
\toprule
\textbf{Direct (DIR)}\par
Place the mustard bottle on the table to the right of the raisin box. \\
\addlinespace
\textcolor{mover}{\textbf{Target-first (TF)}}\par
Move the mustard bottle on the table so that \textbf{the mustard bottle} is to the right of the raisin box. \\
\addlinespace
\textcolor{reference}{\textbf{Reference-first (RF)}}\par
Move the mustard bottle on the table so that \textbf{the raisin box} is to the left of the mustard bottle. \\
\bottomrule
\end{tabular}
\end{minipage}
\end{table}

\section{Paired Spatial Evaluation}
To test whether wording changes behavior without changing the task, we repeat each placement from the same physical start under three instructions. We compare how often the requested arrangement is reached, then examine whether that score captures the intended manipulation and its final outcome.

\begin{samepage}
We use four tabletop scenes from RoboLab, a simulation benchmark for generalist manipulation policies~\citep{yang2026robolab}. The cube scene asks the robot to place a cube left, right, in front of, or behind a bowl. The butter and mustard scenes each ask it to place that object left, right, or on top of a raisin box. The bowl scene asks it to stack either bowl on the other. Seven goals are provided by RoboLab, and five extend the same scenes without changing their objects (Appendix~\ref{app:tasks}).
\par
\end{samepage}

Table~\ref{tab:setup} illustrates the three instruction forms using the same mustard-placement goal. The direct form (DIR) is a concise command that retains RoboLab's original wording where available. Target-first (TF) expands the command by describing the mustard as right of the box. Reference-first (RF) instead describes the box as left of the mustard, while still asking the robot to move the mustard. Comparing TF with RF tests this change in perspective; DIR versus TF compares concise and expanded commands. The DIR versus RF comparison combines both changes and appears in Appendix~\ref{app:behavior-results}. For bowl stacking, ``underneath and supporting'' is less specific than the nesting required by the scorer, so we also analyze the results without these goals.

We evaluate the DROID versions of Cosmos3 Nano, Cosmos3 Edge, and FLUX 3 Action~\citep{nvidia2026nano,nvidia2026edge,bfl2026fluxaction}. Nano and FLUX were selected for strong RoboLab performance, with Edge providing a smaller Cosmos comparison~\citep{robolab2026leaderboard}. Each receives wrist and two exterior camera views plus the robot's proprioceptive state. We vary object poses to obtain eight starts per scene that a scripted controller can solve. For every model and goal, we reset the robot and objects before each instruction, so all three wordings receive identical initial observations. These are our \emph{paired runs}. The resulting 864 episodes last 20--40 seconds, including continued control after placement (Appendix~\ref{app:settings}).

A \emph{stable placement} requires the requested relation to hold for one continuous second while the target is supported, moves slower than 2\,cm/s, and does not touch the gripper. Directions use the robot's coordinate frame. We report two rates: the proportion of runs that achieve a stable placement at any time during the episode, and the proportion that meet the placement conditions throughout the final second. Neither check identifies which object moved, however. A cube can end up behind a bowl because the robot moves the cube backward or the bowl forward.

For each model and goal, we compare the outcomes under different wordings from the same starts. We average over goals within scenes, then give each scene and model equal weight. A positive TF--RF difference therefore means that more runs reach the placement under TF. Uncertainty reflects the shared starts rather than treating all rollouts as independent. Three requested relations hold before the robot acts (Fig.~\ref{fig:goals}). We also report rates excluding these goals (Appendices~\ref{app:statistics} and~\ref{app:counts}).

\section{Results}
\begin{figure}[!ht]
\centering
\includegraphics[width=\linewidth]{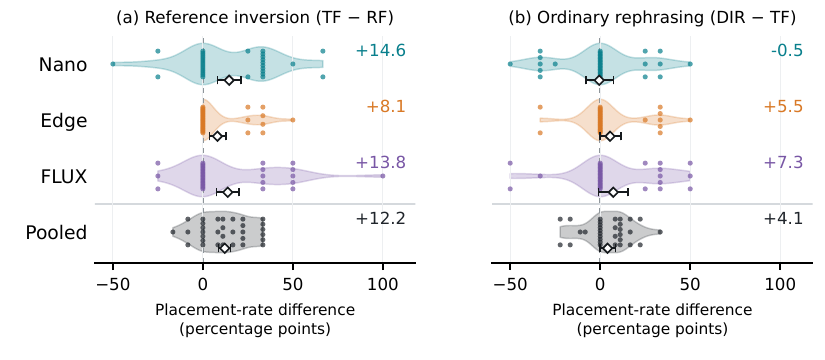}
\caption{Wording effects across 32 starts per row. Each dot's horizontal position shows that start's average placement-rate difference; vertical offsets only separate dots. Wider shading indicates more starts with similar effects. Diamonds and bars show means and 95\% confidence intervals. Positive values favor TF in (a) and DIR in (b).}
\label{fig:wording}
\end{figure}

Task competence under one description does not guarantee consistency across equivalent instructions. Across the three models, the proportion of runs reaching a stable placement is 12.2 percentage points higher under TF than RF (95\% interval 8.6--15.4, adjusted $p<0.001$; Fig.~\ref{fig:wording}). This advantage appears in all three models even though each wording requests the same manipulation from the same physical start. TF and RF do change both object order and relation vocabulary, so this comparison does not identify which aspect drives the effect.

\FloatBarrier

\begin{figure}[!htb]
\centering
\includegraphics[width=\linewidth]{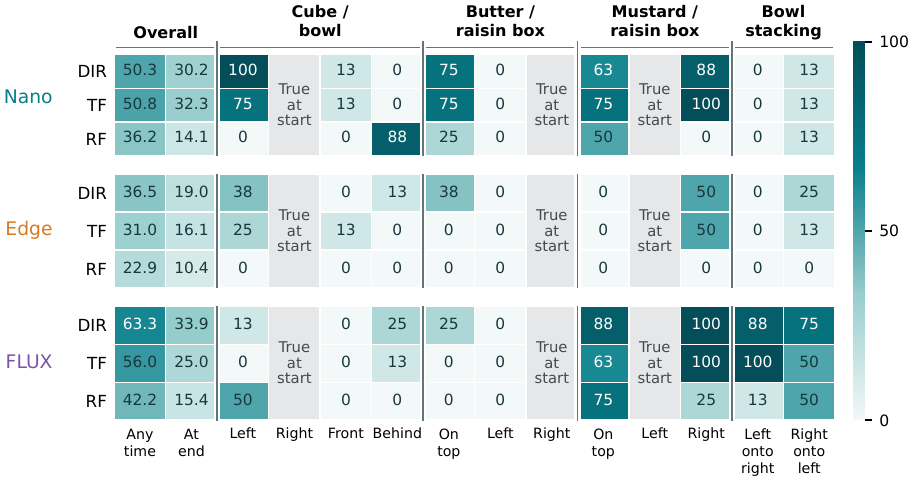}
\caption{Goal cells show the percentage of eight runs reaching a stable placement at any time (100 means all eight). Gray columns mark goals already true at the start. Overall includes these runs and compares rates at any time and at episode end.}
\label{fig:goals}
\end{figure}

The average advantage does not provide a reliable prompting rule for every task. Placing mustard to the right of the box succeeds in 20 of 24 TF runs but only two RF runs. In the cube scene, however, effects in opposite directions cancel: RF performs better for cube-behind despite the scene's zero average gap (Fig.~\ref{fig:goals}). An aggregate score can therefore hide sensitivity to individual relations. The overall TF advantage also persists when the ambiguous bowl-stacking goals are removed (Appendix~\ref{app:statistics}).

A scored arrangement does not necessarily mean the robot followed the instruction. In one Nano rollout, the cube ends up behind the bowl because the robot moves the bowl rather than the requested cube (Appendix~\ref{app:executions}). Arrangements can also be lost: the overall placement rate falls from 43.2\% at any time to 21.8\% at episode end. Evaluation must therefore verify which object moved and whether the requested arrangement was retained.

\FloatBarrier
\section{Interpreting the Requested Goal}
\begin{figure}[!ht]
\centering
\includegraphics[width=\linewidth]{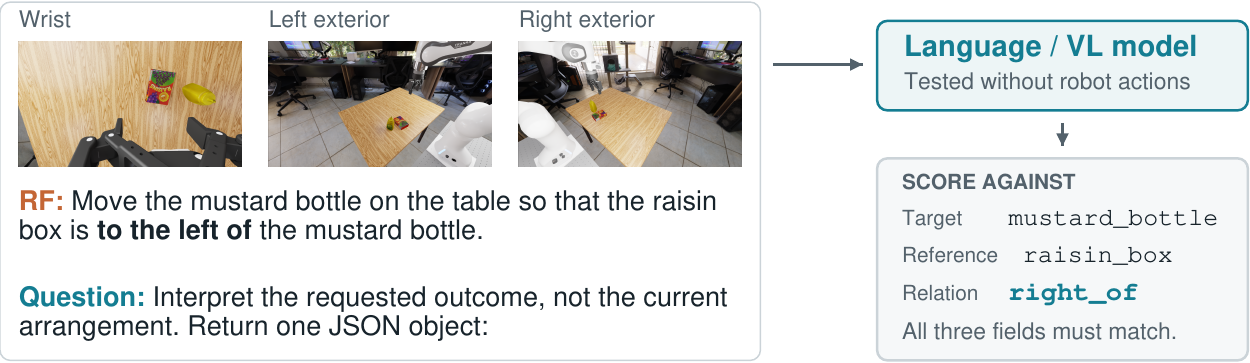}
\caption{Exact RF example and question excerpt. Camera descriptions identify viewpoints and map image directions to the robot frame.}
\label{fig:vqa-protocol}
\end{figure}

To assess instruction understanding independently of manipulation performance, we tested each policy's vision-language backbone on the same ten cube, butter, and mustard goals (Fig.~\ref{fig:vqa-protocol}). We compare text alone with initial images from eight starts per scene (Appendix~\ref{app:vqa}).

\begin{table}[!ht]
\centering\small
\caption{Instruction accuracy under constrained decoding: correct out of ten (text) or equal-scene percentage (images + camera descriptions). $\dagger$ FLUX's image scores test standalone Qwen; the policy uses Qwen for text only.}
\label{tab:vqa}
\begin{tabular}{@{}llcc@{\quad}cc@{}}
\toprule
& & \multicolumn{2}{c}{Text only (/10)} & \multicolumn{2}{c}{With images (\%)}\\
Policy & Language model tested & TF & RF & TF & RF\\
\midrule
Nano & Qwen3-VL-8B-Instruct & 10 & 10 & 100.0 & 89.9\\
Edge & Cosmos3-Edge reasoner & 8 & 4 & 48.3 & 23.3\\
FLUX & Qwen3-VL-4B-Instruct & 10 & 4 & $100.0^\dagger$ & $41.7^\dagger$\\
\bottomrule
\end{tabular}
\end{table}

Reference-first instructions can be misinterpreted even when the language model is evaluated on its own. FLUX processes images separately from its text encoder. Tested alone, the encoder correctly identifies the mustard and box but interprets the RF instruction in Table~\ref{tab:setup} as requiring the mustard to be \emph{left} of the box. It reverses the relation in all six text-only RF errors. Edge also interprets fewer RF than TF instructions correctly (Table~\ref{tab:vqa}).

Adding images can disrupt correct text-only interpretations. Nano answers every text-only instruction correctly, but adding images and camera descriptions introduces errors on two RF instructions: RF accuracy falls to 89.9\% while TF remains correct. Camera descriptions alone change no Nano answers. These policies' backbones interpret equivalent descriptions inconsistently, which may contribute to wording-dependent rollout failures.

\section{Concluding Remarks}
Our results show why instruction consistency should be evaluated alongside task performance. The average TF advantage over RF masks opposing effects across spatial relations, and spatial scores can count placements achieved by moving the wrong object. Evaluations should therefore compare equivalent goal descriptions and verify which object moved and whether the requested arrangement persists. Controlled VLM backbone tests complement these behavioral checks, but do not establish the causes of rollout failures or effects of robot training (Appendix~\ref{app:vqa}).

Future work should extend this evaluation to vision-language-action models (VLAs), comparing them with WAMs under matched tasks, initial states, and scoring criteria. We also aim to develop a benchmark spanning broader scenes, varied spatial descriptions, and explicitly defined robot-, camera-, and object-relative frames of reference.
\clearpage
\bibliography{references}
\clearpage
\appendix

\setlength{\parskip}{5.5pt}
\section{Evaluation details and full instructions}
\label{app:tasks}
The comparison changes the instruction while fixing the requested placement and the object the robot should move. For each physical starting state, the same model is run separately under the direct (DIR), target-first (TF), and reference-first (RF) instructions. ``Target-first'' describes the order of objects in the relational clause after ``so that''. Every instruction still names the same target object as the object to move. The control sequence is not fixed: each policy generates a new sequence from its instruction, so later observations may differ.

\subsection{Scenes and starting states}
The study includes eight starting states in each of four native RoboLab scenes. Small changes to object positions and orientations produce different starts, and a scripted controller checks the requested placements before learned-policy evaluation. The cube scene has four goals, each box scene has three, and the two-bowl scene has two, giving twelve goals and 36 instructions. With three models and eight starts for every goal, this yields 864 episodes. The planned two-bin scene did not produce a qualifying start with the scripted controller and was excluded before the learned-policy outcomes were collected. It contributes no observations to these results.

\begin{table}[htbp]
\centering\small
\caption{Scenes, goals, and episode duration. All goals in a scene share its eight starting states. The already-satisfied column describes all eight starts in that scene.}
\label{tab:appendix-scenes}
\begin{tabular}{@{}p{.25\linewidth}p{.36\linewidth}p{.20\linewidth}r@{}}
\toprule Scene & Requested placements & Already satisfied & Duration\\
\midrule
Cube and bowl & Cube left, right, in front, behind bowl & Cube right & 30\,s\\
Butter and raisin box & Butter left, right, on top of box & Butter right & 40\,s\\
Mustard and raisin box & Mustard left, right, on top of box & Mustard left & 40\,s\\
Two bowls & Either bowl on the other & None & 20\,s\\
\bottomrule
\end{tabular}
\end{table}

Directions are defined in the robot's coordinate frame. The left and right bowl identities are assigned from their positions at reset and remain fixed for the episode. Seven goals are supplied by RoboLab, and five add left/right placements using these same scenes and objects. In particular, the cube-right instruction is tested in the simple cube/bowl/banana scene used for the other cube goals, rather than in RoboLab's separate cube-right scene with additional objects.

\begin{figure}[htbp]
\centering
\includegraphics[width=\linewidth]{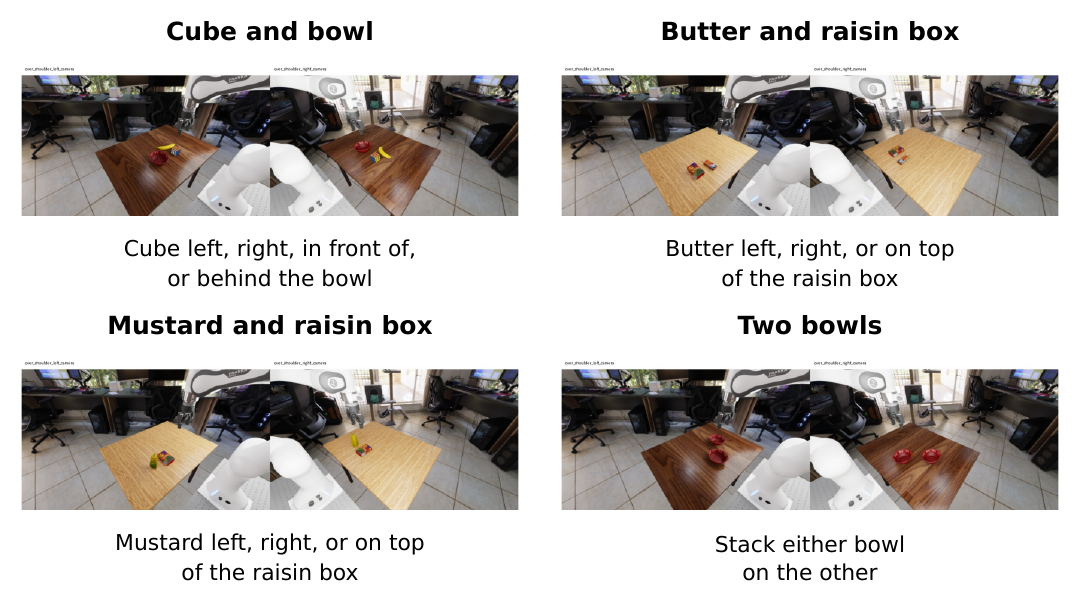}
\caption{Illustrative views of the four RoboLab scenes. Object poses vary across starting states; all wording comparisons within a start use identical physical states.}
\label{fig:scene-overview}
\end{figure}
\FloatBarrier
\subsection{Exact instructions}
The tables below reproduce all tested instructions. DIR retains the benchmark's original wording where the goal is provided for that scene. TF describes the relation from the target object's perspective. RF describes the same placement from the reference object's perspective, using the converse relation. Spelling and capitalization are preserved.

\subsubsection{Cube and bowl}

\begin{longtable}{@{}p{.08\linewidth}p{.87\linewidth}@{}}
\toprule Form & Instruction\\\midrule\endhead

\multicolumn{2}{@{}l}{\textbf{Cube left of bowl}}\\

DIR & Put the rubiks cube to the left of the bowl\\[3pt]

TF & Place the Rubik's cube so that the Rubik's cube is to the left of the bowl.\\[3pt]

RF & Place the Rubik's cube so that the bowl is to the right of the Rubik's cube.\\[3pt]

\addlinespace

\multicolumn{2}{@{}l}{\textbf{Cube right of bowl}}\\

DIR & Put the rubiks cube to the right of the bowl\\[3pt]

TF & Place the Rubik's cube so that the Rubik's cube is to the right of the bowl.\\[3pt]

RF & Place the Rubik's cube so that the bowl is to the left of the Rubik's cube.\\[3pt]

\addlinespace

\multicolumn{2}{@{}l}{\textbf{Cube in front of bowl}}\\

DIR & Put the rubiks cube in front of the bowl\\[3pt]

TF & Place the Rubik's cube so that the Rubik's cube is in front of the bowl.\\[3pt]

RF & Place the Rubik's cube so that the bowl is behind the Rubik's cube.\\[3pt]

\addlinespace

\multicolumn{2}{@{}l}{\textbf{Cube behind bowl}}\\

DIR & Put the rubiks cube behind the bowl\\[3pt]

TF & Place the Rubik's cube so that the Rubik's cube is behind the bowl.\\[3pt]

RF & Place the Rubik's cube so that the bowl is in front of the Rubik's cube.\\[3pt]

\addlinespace

\bottomrule\end{longtable}

\subsubsection{Butter and raisin box}

\begin{longtable}{@{}p{.08\linewidth}p{.87\linewidth}@{}}
\toprule Form & Instruction\\\midrule\endhead

\multicolumn{2}{@{}l}{\textbf{Butter box on raisin box}}\\

DIR & Pick up the butter box and place it on top of the raisin box\\[3pt]

TF & Place the butter box so that the butter box is on top of and supported by the raisin box.\\[3pt]

RF & Place the butter box so that the raisin box is underneath and supporting the butter box.\\[3pt]

\addlinespace

\multicolumn{2}{@{}l}{\textbf{Butter box left of raisin box}}\\

DIR & Place the butter box on the table to the left of the raisin box.\\[3pt]

TF & Move the butter box on the table so that the butter box is to the left of the raisin box.\\[3pt]

RF & Move the butter box on the table so that the raisin box is to the right of the butter box.\\[3pt]

\addlinespace

\multicolumn{2}{@{}l}{\textbf{Butter box right of raisin box}}\\

DIR & Place the butter box on the table to the right of the raisin box.\\[3pt]

TF & Move the butter box on the table so that the butter box is to the right of the raisin box.\\[3pt]

RF & Move the butter box on the table so that the raisin box is to the left of the butter box.\\[3pt]

\addlinespace

\bottomrule\end{longtable}

\clearpage
\subsubsection{Mustard and raisin box}

\begin{longtable}{@{}p{.08\linewidth}p{.87\linewidth}@{}}
\toprule Form & Instruction\\\midrule\endhead

\multicolumn{2}{@{}l}{\textbf{Mustard bottle on raisin box}}\\

DIR & Place the mustard on the raisin box.\\[3pt]

TF & Place the mustard bottle so that the mustard bottle is on top of and supported by the raisin box.\\[3pt]

RF & Place the mustard bottle so that the raisin box is underneath and supporting the mustard bottle.\\[3pt]

\addlinespace

\multicolumn{2}{@{}l}{\textbf{Mustard bottle left of raisin box}}\\

DIR & Place the mustard bottle on the table to the left of the raisin box.\\[3pt]

TF & Move the mustard bottle on the table so that the mustard bottle is to the left of the raisin box.\\[3pt]

RF & Move the mustard bottle on the table so that the raisin box is to the right of the mustard bottle.\\[3pt]

\addlinespace

\multicolumn{2}{@{}l}{\textbf{Mustard bottle right of raisin box}}\\

DIR & Place the mustard bottle on the table to the right of the raisin box.\\[3pt]

TF & Move the mustard bottle on the table so that the mustard bottle is to the right of the raisin box.\\[3pt]

RF & Move the mustard bottle on the table so that the raisin box is to the left of the mustard bottle.\\[3pt]

\addlinespace

\bottomrule\end{longtable}

\subsubsection{Two bowls}

\begin{longtable}{@{}p{.08\linewidth}p{.87\linewidth}@{}}
\toprule Form & Instruction\\\midrule\endhead

\multicolumn{2}{@{}l}{\textbf{Left bowl onto right bowl}}\\

DIR & Stack the left bowl on the right bowl\\[3pt]

TF & Move the left bowl so that the left bowl is stacked on the right bowl.\\[3pt]

RF & Move the left bowl so that the right bowl is underneath and supporting the left bowl.\\[3pt]

\addlinespace

\multicolumn{2}{@{}l}{\textbf{Right bowl onto left bowl}}\\

DIR & Stack the right bowl on the left bowl\\[3pt]

TF & Move the right bowl so that the right bowl is stacked on the left bowl.\\[3pt]

RF & Move the right bowl so that the left bowl is underneath and supporting the right bowl.\\[3pt]

\addlinespace

\bottomrule\end{longtable}

The bowl task deserves separate interpretation. Its DIR instruction requests stacking, and its TF instruction likewise specifies that the target is stacked on the other bowl. RF describes the lower bowl as underneath and supporting the target bowl. That wording does not explicitly require the nesting checked by the benchmark predicate. The bowl results are therefore reported separately and the pooled wording contrast is also examined without them.

\section{Policy settings and evaluation criteria}\label{app:settings}
The policies receive a wrist-camera view, two fixed exterior views, and robot state. RoboLab's camera configuration is held fixed across instruction forms. Each policy produces 32 actions at a time, executed at 15\,Hz before the next observation. The last action sequence ends at the episode horizon. Paired instructions use the same random seed (6100). Episodes continue for the full duration in Table~\ref{tab:appendix-scenes}, including after a successful placement. Each checkpoint retains its released policy implementation and image preprocessing.

\begin{table}[htbp]
\centering\small
\caption{Inference settings used for the three evaluated policies.}
\label{tab:appendix-settings}
\begin{tabular}{@{}lccc@{}}
\toprule Setting & Cosmos3 Nano & Cosmos3 Edge & FLUX 3 Action\\
\midrule
Arithmetic & bfloat16 & bfloat16 & bfloat16\\
Sampling steps & 4 & 4 & 4\\
Sampler & UniPC & UniPC & Cosmos UniPC\\
Noise shift & 5.0 & 5.0 & 5.0\\
Visual guidance & 3.0 & 3.0 & 4.0\\
Action guidance & --- & --- & 1.0\\
Observation history & 1 & 1 & 1\\
Actions per request & 32 & 32 & 32\\
Control frequency & 15\,Hz & 15\,Hz & 15\,Hz\\
Prompt JSON formatting & No & Yes & No\\
Input/canvas setting & $540\!\times\!640$ & $540\!\times\!640$ & $544\!\times\!736$\\
\bottomrule
\end{tabular}
\end{table}

The Cosmos models use joint-position actions and a resolution setting of 480; Edge's guidance interval is [960,1001]. FLUX uses its DROID camera layout and absolute actions with a scale of 2.0. The different input dimensions in Table~\ref{tab:appendix-settings} reflect image preprocessing; the simulated cameras are unchanged. The exact full-policy revision for FLUX was not recorded, which limits checkpoint reproducibility. Its text encoder was verified separately (Appendix~\ref{app:vqa}).

\subsection{Placement criteria and statistical comparison}\label{app:statistics}
For left/right/front/behind, the spatial predicate uses the native 45-degree relation cone in the robot frame and requires the target to be supported by the table. For on-top placement, the native support predicate checks upward support and the target centroid within the reference footprint, with a 1\,cm tolerance. Bowl stacking combines the native open-top containment predicate with contact between the bowls. In every case the gripper must be detached from the target. A stable placement requires the full predicate and target speed below 2\,cm/s for one continuous second. Speed is computed by a backward difference of successive recorded positions at 15\,Hz, which avoids treating solver velocity jitter at an unchanged position as motion.

The any-time outcome asks whether this condition holds anywhere in the episode. The episode-end outcome requires it throughout the final second. Neither condition proves that the intended object was manipulated. For example, moving the reference can make the requested relative arrangement true while leaving the target untouched. They are placement outcomes rather than a complete instruction-following score.

Cube-right, butter-right, and mustard-left are already satisfied in all eight saved starting states. These three goals account for 216 episodes; the remaining nine account for 648. Initially satisfied goals pass the any-time criterion under every instruction form. They therefore increase absolute placement rates while contributing zero to the paired TF--RF difference. We identify these goals from the saved physical states and report rates both with and without them.

For a wording comparison, the two binary placement outcomes are subtracted for the same model, starting state, and goal. For example, if TF reaches the placement and RF does not, the paired difference is +1. The reverse gives $-1$, and agreement gives 0. Differences are averaged over goals within each scene, then equally over scenes and models. This prevents scenes with more goals from dominating the pooled result. Confidence intervals use 10,000 bootstrap draws of the eight physical starts within each scene. All associated goals, models, and instruction forms remain together. For the original pooled any-time TF--RF and DIR--TF tests, we independently flip the signs of the 32 physical-start means, each averaged over the associated models and goals. We use 100,000 draws for two-sided tests and apply Holm correction across the two comparisons. Per-model, subgroup, per-relation, terminal-outcome, and newly added DIR--RF summaries are descriptive. They do not expand that original test family.

The pooled TF--RF estimate is 12.15 percentage points for any-time placement and 11.20 points for episode-end placement. Removing the bowl scene and giving each remaining scene equal weight leaves an any-time estimate of 10.65 points. This descriptive sensitivity check shows that the main direction does not depend on the bowl-language ambiguity. It does not establish equivalence of the bowl prompts. DIR--TF is 4.08 points, with a 95\% interval of $[-0.2,8.4]$ and an adjusted $p=0.089$. Failure to detect that difference is not evidence that instruction expansion has no effect.

\section{Additional behavioral results}
\label{app:behavior-results}
\subsection{Direct versus reference-first wording}
\begin{figure}[htbp]\centering
\includegraphics[width=.90\linewidth]{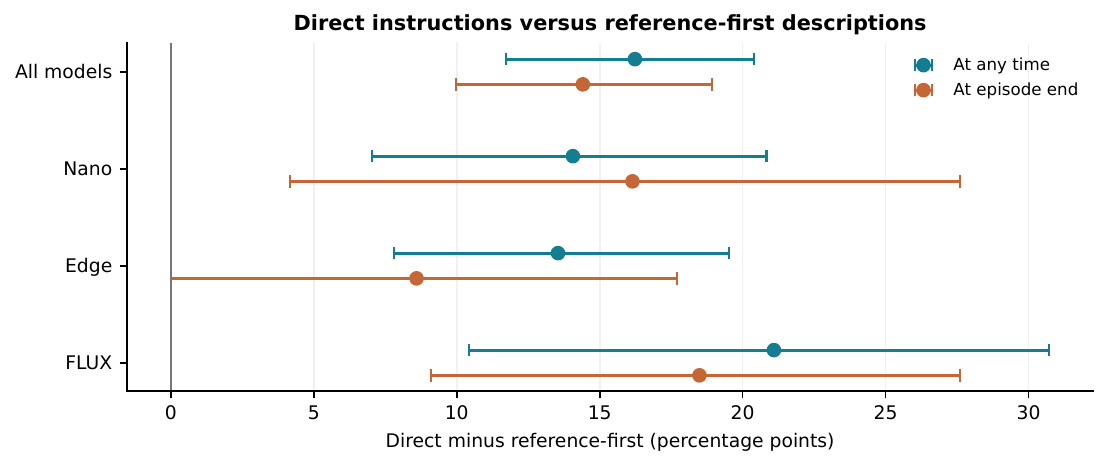}
\caption{The short direct instruction also outperforms reference-first descriptions on average. Intervals are descriptive paired-start bootstrap intervals. DIR--RF was added after the original analyses.}
\label{fig:appendix-di}\end{figure}

The exploratory DIR--RF comparison connects reference-first wording to the short instruction normally supplied with a task. The forms also differ in length and sentence construction, so this comparison cannot isolate the effect of changing spatial perspective. Table~\ref{tab:appendix-di} and Figure~\ref{fig:appendix-di} report descriptive estimates.
\begin{table}[htbp]
\centering\small
\caption{Exploratory DIR--RF gaps in percentage points, with 95\% intervals obtained by resampling paired physical starts. Positive values favor the direct instruction. No new multiplicity-adjusted significance claims are made.}
\label{tab:appendix-di}
\begin{tabular}{@{}lcc@{}}\toprule Model & At any time & At episode end\\\midrule

All models & 16.2 [11.7, 20.4] & 14.4 [10.0, 18.9]\\

Nano & 14.1 [7.0, 20.8] & 16.1 [4.2, 27.6]\\

Edge & 13.5 [7.8, 19.5] & 8.6 [0.0, 17.7]\\

FLUX & 21.1 [10.4, 30.7] & 18.5 [9.1, 27.6]\\

\bottomrule\end{tabular}\end{table}

\subsection{Relation-specific effects}
\begin{figure}[htbp]\centering
\includegraphics[width=\linewidth]{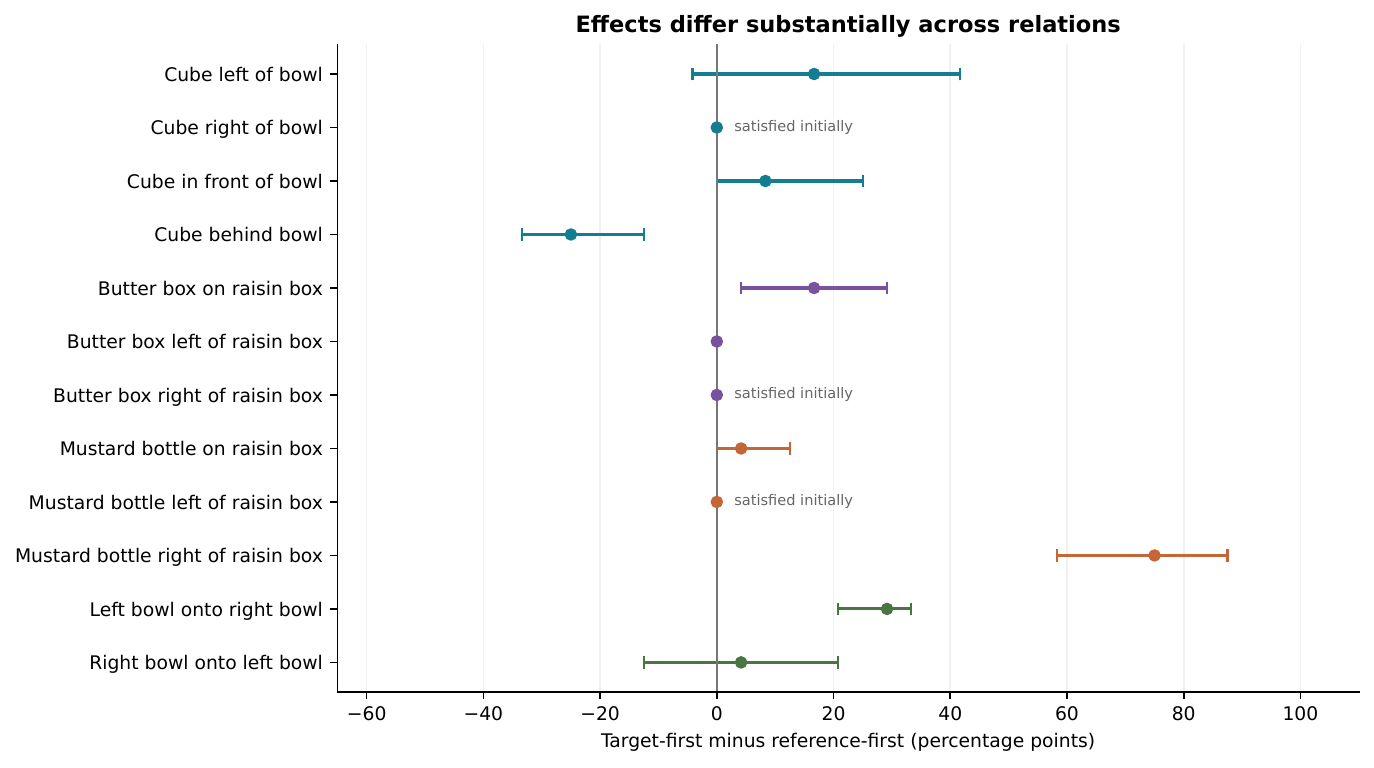}
\caption{Per-relation TF--RF differences pooled over models, with descriptive 95\% intervals from the eight physical starts for that scene. Positive values favor target-first wording. Initially-satisfied goals contribute zero to this any-time contrast.}
\label{fig:appendix-relation}\end{figure}

Figure~\ref{fig:appendix-relation} shows that the average wording effect does not represent a uniform drop across relations. Mustard-right exhibits the largest contrast: 20 of 24 target-first episodes reach the placement, compared with 2 of 24 reference-first episodes. These counts pool the three models over eight starts each. The corresponding target-first/reference-first counts are 8/0 for Nano, 4/0 for Edge, and 8/2 for FLUX. By contrast, cube-behind has a negative TF--RF gap, and the cube scene has no net TF--RF gap after averaging its four relations. This pattern motivates testing equivalent descriptions within each relation rather than expecting one prompt form to be uniformly preferable. Per-relation estimates involve only eight physical starts and should be read as descriptive localization of the observed effect.

\subsection{Placement at any time and at episode end}
\begin{figure}[htbp]\centering
\includegraphics[width=\linewidth]{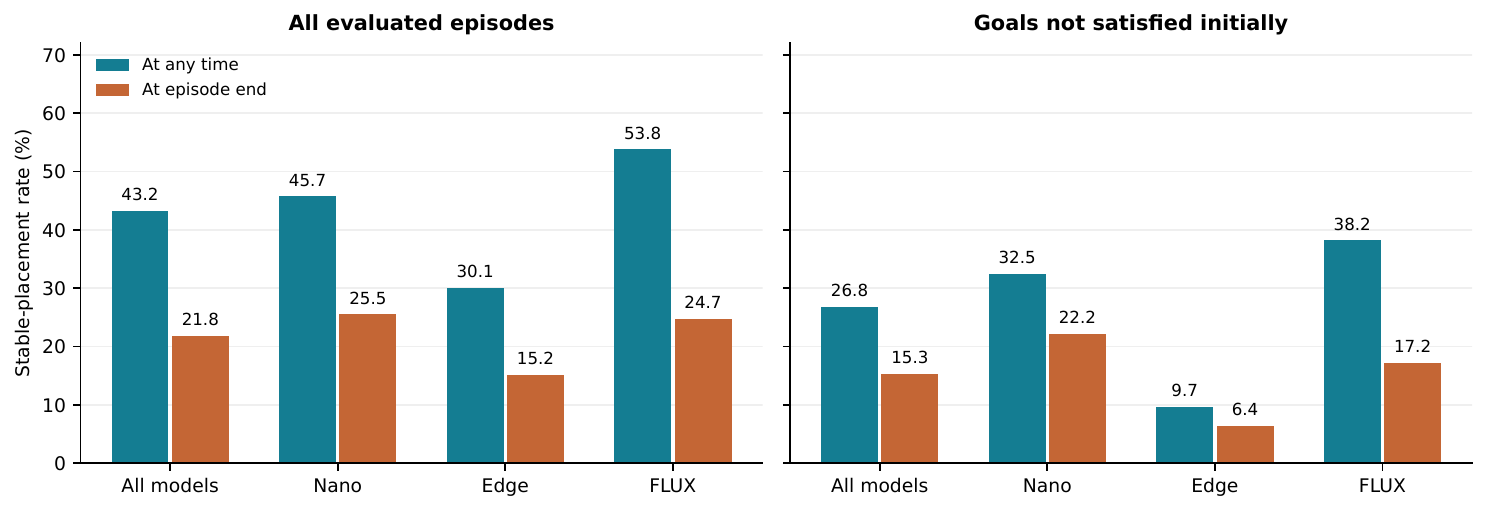}
\caption{Any-time and episode-end placement rates, averaging goals within each scene and weighting scenes equally. The right panel excludes goals already satisfied at reset. The episode-end check requires the placement conditions throughout the final second.}
\label{fig:appendix-end}\end{figure}

Across the 864 episodes, 383 reach a stable placement and 191 also satisfy the episode-end criterion. Thus, 192 of the 383 any-time passes (50.1\%) fail the episode-end check. Among the 648 episodes whose goals are not satisfied initially, 167 reach a stable placement and 94 pass the final check; the remaining 73 account for 43.7\% of that group's any-time passes. For initially-satisfied goals, 97 of 216 episodes pass the final check. An episode can lose and regain a placement, so endpoint success does not establish continuous retention after first attainment. These are raw counts. Figure~\ref{fig:appendix-end} uses the scene-balanced rates reported in the main paper, so its percentages need not equal the raw fractions.

\subsection{Complete per-goal counts}\label{app:counts}
Each entry gives any-time and episode-end stable-placement counts, both out of eight starts. A dagger marks a goal already satisfied in all eight starting states. Zeros indicate observed failures. These scores assess the resulting arrangement and do not establish that the requested object moved.

\begin{table}[htbp]\centering\small
\caption{Nano placement counts. Entries are any-time/end, each out of eight.}
\begin{tabular}{@{}p{.48\linewidth}ccc@{}}\toprule Goal & Direct & Target-first & Reference-first\\\midrule

Cube left of bowl & 8/2 & 6/3 & 0/0\\

Cube right of bowl$^\dagger$ & 8/3 & 8/2 & 8/0\\

Cube in front of bowl & 1/1 & 1/1 & 0/0\\

Cube behind bowl & 0/0 & 0/0 & 7/6\\

Butter box on raisin box & 6/5 & 6/5 & 2/1\\

Butter box left of raisin box & 0/0 & 0/0 & 0/0\\

Butter box right of raisin box$^\dagger$ & 8/5 & 8/4 & 8/6\\

Mustard bottle on raisin box & 5/5 & 6/5 & 4/2\\

Mustard bottle left of raisin box$^\dagger$ & 8/5 & 8/5 & 8/0\\

Mustard bottle right of raisin box & 7/3 & 8/6 & 0/0\\

Left bowl onto right bowl & 0/0 & 0/0 & 0/0\\

Right bowl onto left bowl & 1/1 & 1/1 & 1/0\\

\bottomrule\end{tabular}\end{table}

\begin{table}[htbp]\centering\small
\caption{Edge placement counts. Entries are any-time/end, each out of eight.}
\begin{tabular}{@{}p{.48\linewidth}ccc@{}}\toprule Goal & Direct & Target-first & Reference-first\\\midrule

Cube left of bowl & 3/1 & 2/1 & 0/0\\

Cube right of bowl$^\dagger$ & 8/4 & 8/3 & 8/0\\

Cube in front of bowl & 0/0 & 1/0 & 0/0\\

Cube behind bowl & 1/0 & 0/0 & 0/0\\

Butter box on raisin box & 3/2 & 0/0 & 0/0\\

Butter box left of raisin box & 0/0 & 0/0 & 0/0\\

Butter box right of raisin box$^\dagger$ & 8/5 & 8/3 & 8/7\\

Mustard bottle on raisin box & 0/0 & 0/0 & 0/0\\

Mustard bottle left of raisin box$^\dagger$ & 8/3 & 8/4 & 8/3\\

Mustard bottle right of raisin box & 4/3 & 4/4 & 0/0\\

Left bowl onto right bowl & 0/0 & 0/0 & 0/0\\

Right bowl onto left bowl & 2/1 & 1/1 & 0/0\\

\bottomrule\end{tabular}\end{table}

\begin{table}[htbp]\centering\small
\caption{FLUX placement counts. Entries are any-time/end, each out of eight.}
\begin{tabular}{@{}p{.48\linewidth}ccc@{}}\toprule Goal & Direct & Target-first & Reference-first\\\midrule

Cube left of bowl & 1/0 & 0/0 & 4/3\\

Cube right of bowl$^\dagger$ & 8/2 & 8/2 & 8/8\\

Cube in front of bowl & 0/0 & 0/0 & 0/0\\

Cube behind bowl & 2/0 & 1/0 & 0/0\\

Butter box on raisin box & 2/2 & 0/0 & 0/0\\

Butter box left of raisin box & 0/0 & 0/0 & 0/0\\

Butter box right of raisin box$^\dagger$ & 8/4 & 8/6 & 8/3\\

Mustard bottle on raisin box & 7/2 & 5/2 & 6/2\\

Mustard bottle left of raisin box$^\dagger$ & 8/6 & 8/4 & 8/0\\

Mustard bottle right of raisin box & 8/8 & 8/3 & 2/0\\

Left bowl onto right bowl & 7/6 & 8/5 & 1/1\\

Right bowl onto left bowl & 6/0 & 4/0 & 4/0\\

\bottomrule\end{tabular}\end{table}

\FloatBarrier
\section{Policy rollout examples and image context}
\label{app:executions}
These policy rollouts illustrate the wording effect and a limitation of spatial scoring. The mustard-right pair shows different outcomes from the same initial state; the cube-behind pair shows a scored relation reached by moving the wrong object. Figure~\ref{fig:full-views} provides the full camera context for both pairs. These examples were selected from the recorded outcome differences before inspecting their videos. They illustrate possible behaviors and are not used to estimate their prevalence.

\subsection{An arrangement can be reached by moving the reference object}
Both Nano instructions ask the robot to move the cube behind the bowl. In the reference-first rollout, however, the robot holds the bowl while the cube remains on the tabletop. Moving the bowl changes the relative positions enough for the cube-behind check to pass. The target-first rollout does not pass that check. Thus, the reference-first score advantage in this example is not evidence of better instruction following: the predicate credits a relation that became true through manipulation of the wrong object.

\begin{figure}[!ht]
\centering
\includegraphics[width=\linewidth]{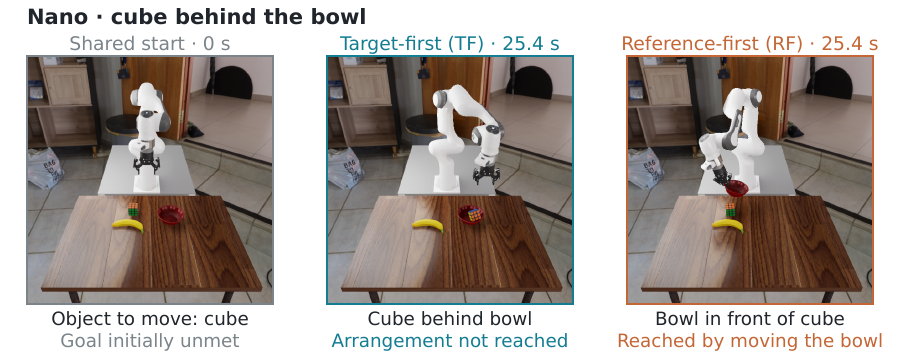}
\caption{Nano's cube-behind policy rollouts. Both outcome snapshots are at 25.4 seconds, when the reference-first run first completes the stable-placement interval. The robot holds the bowl in RF while the cube is visible on the table. This arrangement no longer satisfies the episode-end check. The labels describe recorded spatial outcomes, not complete instruction success.}
\label{fig:wrong-object}
\end{figure}

\subsection{Camera context and matched snapshots}
The mustard example compares Edge's two instruction forms at the end of the 40-second episode. The cube example compares Nano's forms at 25.4 seconds, when the RF rollout first completes the stable-placement interval. Within each pair, both rollouts begin from the same physical state and are shown at the same elapsed time.

\begin{figure}[p]
\centering
\includegraphics[width=\linewidth]{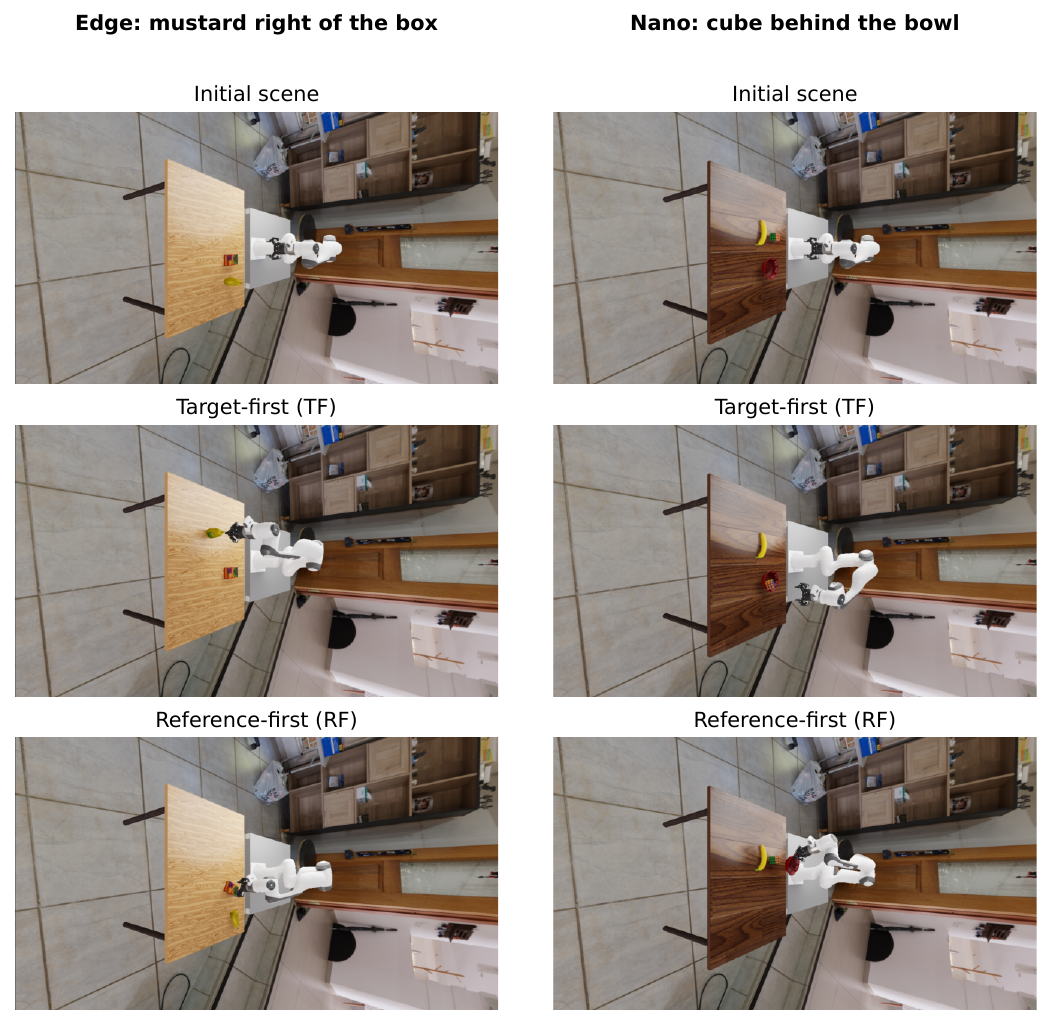}
\caption{Full exterior-camera frames for the two policy rollout examples. Columns correspond to Edge mustard-right and Nano cube-behind. Rows show the shared initial scene, TF, and RF. The camera image is shown in its native orientation, so image left/right should not be interpreted as the robot-frame relation. These are executed observations, not model-generated predictions.}
\label{fig:full-views}
\end{figure}

The close views use a fixed crop of the exterior camera, rotated 90 degrees for readability. The crop is identical across instruction forms, and no image content is retouched. Figure~\ref{fig:full-views} shows the full frames in their original orientation.

\clearpage
\section{Can predicted futures help explain a policy failure?}
\label{app:forecasts}
\suppressfloats[t]

We also asked whether a policy's generated video helps identify an impending manipulation error beyond what is already apparent from the current scene and proposed actions. This was the study's original primary question. The planned secondary analysis of wording became the paper's focus after the behavioral outcomes were available. At that point, we replaced the planned human video annotations with two vision--language models (VLMs), before inspecting their answers. The resulting analysis found no measurable benefit from the video labels, but unreliable annotations prevent a conclusion about the information in the videos themselves.

\subsection{Forecast selection and annotation}

We analyzed one forecast from each of the 864 episodes. Without viewing the forecast or later outcome, we selected the first prediction made when the gripper was within 10\,cm of a movable task object. Two Nano episodes never met this condition; we used their first predictions. Each forecast accompanied up to 32 actions at 15\,Hz. We retained only frames assigned to the actions that the policy subsequently performed, before its next prediction.

Qwen3-VL-235B-A22B-Instruct-FP8 and GLM-4.5V-FP8 independently labeled the moving object, motion direction, final spatial relation, its comparison object (not necessarily the instruction's reference), possible release, and missing or hallucinated objects. They received eight forecast frames in a common three-camera layout and an episode-start image with numbered objects and direction arrows. The instruction, requested goal, policy identity, wording form, policy rollout, and outcome were hidden. Ambiguous observations could be labeled unknown. For disagreements, Qwen selected between the two answers and an unknown option, with answer order randomized. Because Qwen also supplied an original annotation, this step does not provide independent validation.

\subsection{How reliably could the videos be interpreted?}

The annotators agreed on the moving object in 59.1\% of forecasts, but on the final relation in only 20.1\% (Table~\ref{tab:forecast-agreement}). They agreed on every field in 5.9\%. High agreement on rare events, such as release, largely reflects their rarity: chance-corrected agreement on release was near zero. These measures describe agreement, not accuracy.

\begin{table}[htbp]
\centering
\small
\begin{tabular}{lrr}
\toprule
Annotation & Agreement (\%) & Cohen's $\kappa$ \\
\midrule
Moving object & 59.1 & 0.398 \\
Motion direction & 48.6 & 0.108 \\
Final spatial relation & 20.1 & 0.077 \\
Comparison object & 33.3 & 0.111 \\
Possible release & 93.4 & 0.026 \\
Missing object & 97.7 & 0.000 \\
Hallucinated object & 98.5 & 0.231 \\
\bottomrule
\end{tabular}
\caption{Agreement between the two independent video annotators on 864 selected forecasts. The 862 additional episode-start forecasts also showed weak agreement (all $\kappa<0.115$). Neither annotation set has been validated against human judgments.}
\label{tab:forecast-agreement}
\end{table}

After resolving disagreements, we classified the labels relative to the task. A requested or alternative relation was credited only when the labels identified the intended target as moving and matched that goal's comparison object and relation. This gave 4.5\% requested-relation labels, 11.5\% another registered relation, 70.4\% neither, and 13.7\% unknown. ``Neither'' includes forecasts naming a different moving object. Separately, 77.9\% indicated upward-only motion. These model-generated labels are unverified. Short grasping or lifting clips may reveal little about the eventual placement, but annotation errors provide another explanation.

Inspection of two examples exposed overlapping object numbers and direction arrows, occasional object-name confusions, and image stretching in the common layout. The numbered scene image could also precede the selected forecast by several actions. These ambiguities may affect interpretation. We verified that the two inspected predictions matched the corresponding actions and camera, but have not extended that check to every forecast or established that generated motion follows the same timing as the policy rollout.

\subsection{Does the video add diagnostic information?}

We predicted whether each action chunk contained either of two detected errors. Wrong-object manipulation required at least 0.2 seconds of continuous gripper contact with a non-target object and displacement of at least 2\,cm from its position at chunk start. Contradictory release required target detachment after contact, followed by at least 0.2 seconds supported and detached in exactly one registered alternative relation while the requested relation was false, all within the chunk. Alternatives were the other tested cube, butter, and mustard relations; bowls had none. These checks differ from the one-second stable-placement endpoint. The scorer marked 47 of 864 chunks positive: 14 wrong-object and 33 contradictory-release detections. These have not all been visually verified. Contact followed by detachment can count as release without substantive prior manipulation.

The baseline was an L2-regularized logistic classifier using the model, scene, goal, current object and gripper states, distance to the goal, and whether the goal already held. It also used object proximity along the proposed actions' forward kinematics and prior contact and lift indicators. Wording, future states, and final task success were excluded. The augmented classifier added the video annotations, including unknown categories. Both classifiers used fixed regularization ($C=1$) and eight folds, each holding out the same initial-state slot across all conditions. Preprocessing was fitted within the training fold. This comparison uses privileged simulator information and is not a deployable failure detector.

We report the reduction in held-out Brier error, $\Delta B = B_{\text{baseline}}-B_{\text{with video}}$, weighting models equally, scenes equally within models, and goals equally within scenes. Adding video labels slightly worsened the pooled score (Table~\ref{tab:forecast-utility}). The 95\% interval includes zero, and the small positive FLUX estimate does not establish a model-specific benefit.

\begin{table}[htbp]
\centering
\small
\begin{tabular}{lrrrr}
\toprule
Model & Events / forecasts & Baseline & With video & $\Delta B$ \\
\midrule
All models & 47 / 864 & 0.04012 & 0.04085 & $-0.00073$ \\
Cosmos3 Nano & 15 / 288 & 0.04012 & 0.04151 & $-0.00139$ \\
Cosmos3 Edge & 8 / 288 & 0.02491 & 0.02530 & $-0.00040$ \\
FLUX 3 Action & 24 / 288 & 0.06210 & 0.06104 & $+0.00106$ \\
\bottomrule
\end{tabular}
\caption{Held-out Brier error for predicting detected manipulation errors; lower is better. Positive $\Delta B$ favors video annotations. The pooled 95\% interval is $[-0.00177,+0.00016]$, resampling starts within scenes with fitted predictions held fixed. It excludes annotation uncertainty and variation from refitting. Model-specific rows use classifiers refitted on that policy's episodes and are descriptive; they do not decompose the pooled result.}
\label{tab:forecast-utility}
\end{table}

Controls using only indicators of missing annotations or synthetic no-motion labels with the current-goal flag gave negligible changes ($+0.000011$ and $+0.000009$). The latter control did not pass repeated current images through the video annotators. Shuffling forecast associations across starts over 100 seeds gave a mean change of $-0.00053$, with a 2.5th--97.5th percentile range of $[-0.00177,+0.00049]$. This is a permutation distribution, not a confidence interval for the observed effect.

An exploratory comparison found more detected errors when the annotations named the reference object as moving (23/247, 9.3\%) than when they named the requested object (19/540, 3.5\%). This comparison was made after seeing the results and does not adjust for scene, goal, or model. It therefore does not establish added predictive value.

The analysis cannot distinguish an uninformative short forecast from a failure to extract useful information from it. Human validation of object identity, motion, spatial relations, and detected manipulation errors is needed before attributing this null result to the generated videos. The behavioral findings in the main paper do not depend on these video annotations.

\clearpage
\section{Matched instruction-understanding diagnostic}
\label{app:vqa}

\paragraph{Purpose and scope.}
These tests separate interpreting an instruction from judging the visible arrangement and generating actions. A preliminary evaluation's prespecified primary test asked whether an image matched the instruction. Nearly constant negative answers made it inconclusive. After seeing those results, we fixed a follow-up protocol with constrained answer formatting, separate image and camera-description conditions, and a revised matching question. The follow-up is exploratory, rather than an independent confirmatory test.

We retain the ten cube, butter, and mustard goals and their 30 exact DIR, TF, and RF instructions. Bowl stacking is excluded because of the support/nesting ambiguity. The visual material contains 24 initial and 48 saved-rollout image sets, each with three simultaneous camera views. All share 24 physical starts across three scenes; additional frames do not add independent scenes. We use existing checkpoints and observations, without training or new policy rollouts. Figure~\ref{fig:vqa-protocol} shows one initial-state example with the exact instruction and a question excerpt.

\paragraph{Tested language models.}
Nano's language component has the same weights as Qwen3-VL-8B-Instruct; Edge's matches the released Cosmos3-Edge reasoner. Identical components are evaluated once. FLUX's frozen encoder matches Qwen3-VL-4B-Instruct, but its policy uses this encoder only for text and processes images separately. Qwen's image results are therefore auxiliary. A different Cosmos3-Nano base reasoner examined in the preliminary test is not a matched predecessor of the evaluated policy. These comparisons do not show degradation of component weights during robot adaptation. Without matched training data, they cannot isolate training-recipe or co-training effects on the full policies.

\paragraph{Questions and scoring.}
The model identifies the object to move, the reference object, and the target's requested final relation. All three must be correct. Constrained decoding enforces a fixed JSON format and the supplied object and relation vocabularies, while allowing incorrect choices. This assisted test separates interpretation errors from formatting failures; its scores are not directly comparable to the preliminary unrestricted answers.

We compare instruction text alone, text with camera-orientation descriptions, text with initial images, and text with both. Image contrasts hold the written prompt fixed. The correct answer follows from the instruction, regardless of the visible arrangement, so scoring instruction interpretation does not require visual-answerability judgments.

Each instruction is answered once per no-image condition, giving ten answers per form. Image accuracy averages starts within goals, goals within scenes, and the three scenes equally. Image-versus-text contrasts compare each image-conditioned answer with its fixed no-image answer, using the same scene weighting. We obtain 95\% confidence intervals from 10,000 bootstrap resamples of physical starts within scenes, keeping conditions paired. These intervals condition on the text answers, scenes, and wording templates; they do not estimate uncertainty over new instructions.

\begin{table}[htbp]
\centering\small
\caption{Complete instruction-understanding results. Entries give correct answers out of ten (text) or scene-balanced accuracy (images). Daggers mark auxiliary FLUX image tests.}
\label{tab:vqa-full}
\begin{tabular}{@{}lrrr@{\qquad}rrr@{}}
\toprule
& \multicolumn{3}{c}{\shortstack{Text only\\(correct out of 10)}}
& \multicolumn{3}{c}{\shortstack{Images and camera descriptions\\(accuracy, \%)}}\\
Component & DIR & TF & RF & DIR & TF & RF\\
\midrule
Nano & 10 & 10 & 10 & 100.0 & 100.0 & 89.9\\
Edge & 6 & 8 & 4 & 30.6 & 48.3 & 23.3\\
FLUX & 10 & 10 & 4 & $100.0^\dagger$ & $100.0^\dagger$ & $41.7^\dagger$\\
\bottomrule
\end{tabular}
\end{table}

\paragraph{Interpretation errors and the effect of images.}
Table~\ref{tab:vqa-full} shows that Edge and FLUX interpret fewer RF than TF instructions correctly from text alone. Every one of FLUX's six text-only RF errors reverses the requested relation. Edge also makes errors on DIR and TF instructions, so its difficulty is not restricted to reference-first wording.

Nano interprets all 30 instructions correctly both with text alone and with camera descriptions but no images. When images and descriptions are supplied together, it makes eight relation-reversal errors: five starts for butter-right and three for cube-behind. Its RF accuracy falls by 10.1 percentage points relative to the same written prompt without images (95\% interval 5.2--14.6). Images without camera descriptions produce four butter-right reversals and one butter-on-top error that substitutes a stacking label for the requested support relation. The RF decrease relative to text alone is 6.9 points (1.4--12.5). Thus, image-conditioned errors are concentrated in particular instructions. Camera descriptions alone change no Nano answers, while their additional effect when images are present remains unresolved: $-3.1$ points, with an interval of $-8.0$ to $+1.7$.

\begin{table}[htbp]
\centering\small
\caption{Camera-description, image, and reference-clause controls. Text columns count correct instructions; image columns give scene-balanced accuracy. Daggers mark auxiliary FLUX image tests.}
\label{tab:vqa-controls}
\begin{tabular}{@{}lrrr@{\qquad}rrr@{}}
\toprule
& \multicolumn{3}{c}{\shortstack{Text with camera descriptions\\(correct out of 10)}}
& \multicolumn{3}{c}{\shortstack{Images without descriptions\\(accuracy, \%)}}\\
Component & DIR & TF & RF & DIR & TF & RF\\
\midrule
Nano & 10 & 10 & 10 & 100.0 & 100.0 & 93.1\\
Edge & 5 & 10 & 4 & 38.9 & 46.2 & 23.3\\
FLUX & 10 & 10 & 3 & $97.2^\dagger$ & $100.0^\dagger$ & $41.7^\dagger$\\
\bottomrule
\end{tabular}\par\smallskip
\begin{tabular}{@{}lcc@{}}
\toprule
\multicolumn{3}{c}{Reference-clause placement controls}\\
Component & \shortstack{Text only\\(correct out of 16)}
& \shortstack{Images and camera descriptions\\(accuracy, \%)}\\
\midrule
Nano & 16 & 100.0\\
Edge & 13 & 50.5\\
FLUX & 16 & $100.0^\dagger$\\
\bottomrule
\end{tabular}
\end{table}

\paragraph{Reference-clause placement.}
Sixteen placement controls cover the eight left/right and front/behind goals. Each pair moves the reference phrase without reversing the direction: ``Relative to the bowl, place the Rubik's cube to the left.'' versus ``Place the Rubik's cube to the left, relative to the bowl.'' Nano and FLUX answer all controls correctly in both conditions; FLUX's image tests remain auxiliary. Edge answers 5/8 reference-early and 8/8 reference-late instructions correctly from text alone. With images and camera descriptions, its reference-early accuracy is 17.7 percentage points lower than reference-late (95\% interval 10.4--25.0). These controls use different sentences from TF/RF and do not settle general noun-order effects.

\paragraph{Judging the visible arrangement.}
The revised question asks whether the current views show the requested relation and permits an insufficient-evidence answer. Labels use saved simulator geometry and contact conditions, retaining the preliminary visibility, spatial-boundary, and gripper-contact exclusions. We query each item under both assignments of the match/nonmatch answer codes and average correctness across them. Answers remain nearly constant negative. Across components and the separately scored initial and saved-rollout images, balanced accuracy ranges from 48.5\% to 50.1\%. It weights match and nonmatch recall equally, so always answering nonmatch gives 50\%. At most one of eleven same-start, same-goal pairs with a changing relation is answered correctly for both frames. This inconclusive test establishes neither wording robustness nor general spatial incompetence.

Judging a visible relation requires an answerable image. Machine review of 36 of 72 image sets, without human validation, found later-rollout images less consistently answerable than initial images. Retaining only originally eligible items deemed answerable by every available reviewer, and excluding unreviewed items, leaves balanced accuracy near the 50\% always-nonmatch baseline. These question-answering tests do not reveal the policy's internal goal representation or establish that a wrong answer causes a rollout failure.

\end{document}